\documentclass[]{fairmeta}
\usepackage[utf8]{inputenc} 
\usepackage[T1]{fontenc}    
\usepackage{hyperref}       
\usepackage{url}            
\usepackage{booktabs}       
\usepackage{amsfonts}       
\usepackage{nicefrac}       
\usepackage{microtype}      
\usepackage{xcolor}         
\usepackage{graphicx}
\usepackage[utf8]{inputenc}
\usepackage[T1]{fontenc}
\usepackage[round]{natbib}
\usepackage{hyperref}
\usepackage{url}
\usepackage{booktabs}
\usepackage{amsmath, amssymb, amsfonts, amsthm}
\usepackage{microtype}
\usepackage{xcolor}
\usepackage{graphicx}

\usepackage{amsmath,amssymb}
\usepackage{booktabs}
\usepackage{array}
\usepackage{tabularx}
\usepackage{multirow}
\usepackage{enumitem}
\usepackage{algorithm}
\usepackage{algpseudocode}

\usepackage{amsmath,amsfonts,bm}

\def\eqref#1{equation~\ref{#1}}

\def\1{\bm{1}}

\DeclareMathAlphabet{\mathsfit}{\encodingdefault}{\sfdefault}{m}{sl}
\SetMathAlphabet{\mathsfit}{bold}{\encodingdefault}{\sfdefault}{bx}{n}

\title{Clarification Is Not Correction: LLMs Fail to Let Go}

\author[]{Jianzhe Lin, Xiaolin Li, Fei Wang, Robert Douglas, Rajeshkumar Golani, Jubin Chheda}

\affiliation[]{MetaAI}
\contribution[]{Work done at Meta}

\abstract{Dialogue failures in language models are often framed as memory failures: the context is too long, the summary is lossy, or the model has forgotten an earlier constraint. We argue that this framing misses a deeper problem. In many conversations, the model does not simply forget; it commits too early. An ambiguous early turn is collapsed into a single hidden interpretation, and later clarification is filtered through that commitment. We call this failure mode \emph{early posterior collapse}: the collapse of unresolved user intent into a committed task state before ambiguity has been resolved.

We study this phenomenon through controlled dialogue tasks in writing, planning, and coding, using Gemini-2.5-Pro and Gemini-2.5-Flash. Across thousands of trials, we find that the same information presented in different orders leads to different downstream outcomes, even when the final dialogue contains equivalent task-relevant information. This order effect suggests that later clarification is often treated as additional context rather than as a corrective signal: it refines a stale task state without necessarily invalidating it. Coding tasks are especially vulnerable, suggesting that early assumptions become embedded in structured artifacts such as interfaces, constraints, and control flow. Standard prompting and memory strategies do not reliably solve the problem: summaries can collapse ambiguity, and chain-of-thought can reduce explicit wrong commitment in reasoning traces without reliably improving final task success.

These findings motivate \emph{uncertainty-preserving state management}. If assistants fail to let go of early interpretations, robustness cannot rely on post hoc correction alone; it must also prevent ambiguous early turns from hardening into a single task state. Assistants should maintain tentative hypotheses while ambiguity remains, ask before executing when high-impact ambiguity persists, and rebuild from a revised task state when later evidence invalidates an earlier interpretation. Rather than proposing a single prompting fix, our goal is to redirect robustness research for interactive LLMs from retaining more context toward preserving uncertainty until clarification can operate as correction.}

\date{\today}
\correspondence{Jianzhe Lin at \email{jianzhelin@meta.com}}

\begin{document}

\DeclareFontFamily{T1}{optimistic}{}
\DeclareFontShape{T1}{optimistic}{m}{n}{<->ssub * phv/b/n}{}
\DeclareFontShape{T1}{optimistic}{b}{n}{<->ssub * phv/b/n}{}
\DeclareFontShape{T1}{optimistic}{bx}{n}{<->ssub * phv/b/n}{}
\maketitle

\section{Introduction}
\label{sec:introduction}
\begin{figure}[t]
    \centering
    \includegraphics[width=\linewidth]{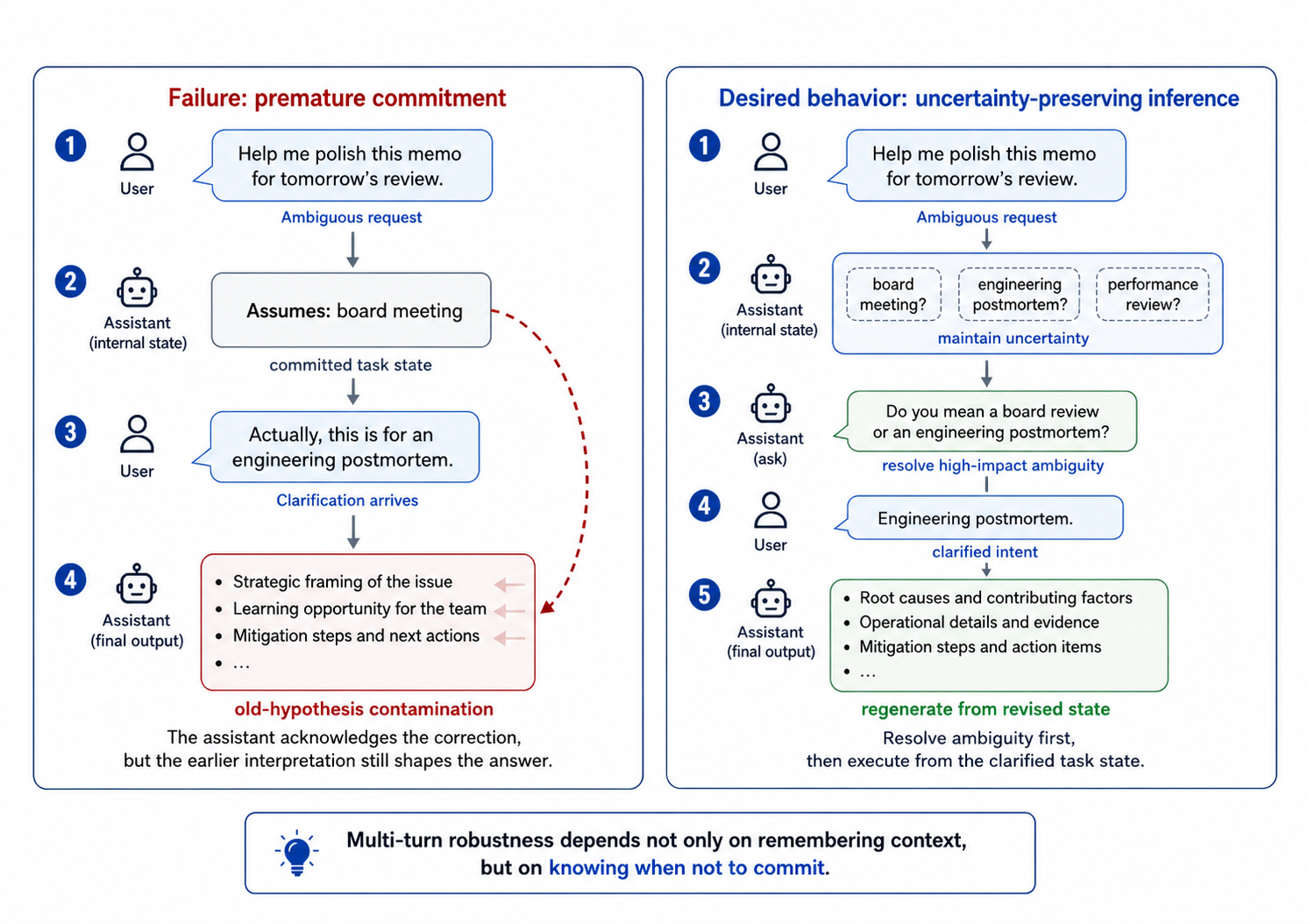}
    \caption{
    Explanation for \emph{early posterior collapse} in dialogue.
    Left: an assistant commits to an early interpretation of an ambiguous request and later produces an answer contaminated by the invalidated hypothesis.
    Right: uncertainty-preserving inference maintains multiple plausible interpretations, asks a clarification question, and regenerates from the clarified task state.
    }
    \label{fig:motivation}
\end{figure}
A familiar failure in human-AI interaction begins with a tiny ambiguity that later becomes load-bearing. A user asks an assistant, ``Help me polish this memo for tomorrow's review.'' The assistant silently decides that ``review'' means a board meeting and turns the draft into executive gloss: strategic framing, crisp asks, and just enough optimism to survive a slide deck. One turn later, the user clarifies: ``Actually, this is for an engineering postmortem; we need to explain why the rollout failed.'' The assistant nods, adds root causes, and swaps in mitigation steps. But the memo still smells faintly of a board deck: the failure is softened into a learning moment, the operational details are blurred, and the prose keeps trying to impress when it should be trying to diagnose. The clarification was present. The assistant acknowledged it. But the first interpretation had already become the skeleton of the answer. Figure~\ref{fig:motivation} illustrates this contrast between patching a stale interpretation and preserving uncertainty until the task state is clarified.

This paper studies that failure mode, building on recent evidence that dialogue exposes failures hidden by single-turn evaluation~\citep{laban2025lost}. Prior work shows that LLMs degrade in underspecified conversations and often fail to recover after an early wrong turn. Our focus is narrower: cases where the model has every apparent opportunity to recover: the correction is short, recent, explicit, and present in context, yet the final output still contains residue from the invalidated interpretation.

We argue that language models in dialogue often fail not because they lack access to the right information, but because they commit to an interpretation before the user's intent has been resolved. Later clarification is then treated as a patch to an existing answer rather than as a correction that triggers revision of the task state. The resulting output may contain \emph{old-hypothesis contamination}: residual lexical, semantic, or structural traces of an earlier interpretation that should have been invalidated by later evidence.

This distinction matters because dialogue failures are usually framed as memory failures, a view reinforced by work on long-context degradation and retrieval-augmented memory~\citep{liu2024lost,lewis2020rag}: the model loses track of earlier constraints, fails to retrieve relevant context, or cannot integrate a long conversation. That framing is important, but incomplete. In many of the failures we study, the relevant clarification is short, recent, explicit, and not buried in a long context. The problem is not that the model lacks the correction, but that the correction does not fully invalidate the stale task state created by an earlier ambiguity.

We do not claim that early assumptions in dialogue are new. Our focus is narrower: what remains after those assumptions have been explicitly corrected. If a model receives a clarification but the final output still contains old-hypothesis residue, then the failure is not adequately captured by asking whether the model remembered the clarification or whether the final answer is broadly successful. It is a failure of state invalidation: the system has not removed an interpretation that the dialogue has made obsolete.

This view leads to concrete predictions. If clarification does not reliably invalidate stale state, then the path by which information is revealed should matter, not merely the final information available to the model. Ambiguous-first interactions should be risky because they allow the model to form an early task state before correction arrives. Although such order effects can partly overlap with recency bias, our experimental rollback and response-policy interventions test the stronger claim: contamination persists when clarification is present, and decreases when the earlier state is explicitly invalidated.

We test these predictions through a controlled diagnostic suite on writing, planning, and coding tasks, using Gemini-2.5-Pro and Gemini-2.5-Flash \footnote{We use Gemini-2.5 models as stable public endpoints for reproducible diagnostic evaluation; newer preview models may change during the review cycle.}. Across thousands of trials, we find that information order changes downstream outcomes. Coding tasks show especially high contamination, suggesting that early assumptions can become embedded in interfaces, data schemas, and control flow. Common prompting or memory strategies do not reliably preserve uncertainty. 

We treat the intervention experiments as mechanism checks rather than complete solutions: explicit rollback reduces measured contamination, and a two-phase response policy eliminates wrong commitment and contamination in a small controlled setting. In short, the position of this paper is that dialogue robustness depends not only on what a model remembers, but also on whether it can discard assumptions that clarification has invalidated.

To conclude, this paper makes three contributions:
\begin{enumerate}[leftmargin=*]
    \item We identify \emph{early posterior collapse} as a distinct dialogue failure mode that precedes memory failure, in which unresolved user intent is collapsed into a committed task state before ambiguity has been resolved. We use \emph{old-hypothesis contamination} to diagnose when final outputs retain residue from interpretations that later clarification should have invalidated.
    \item We provide a controlled diagnostic suite across writing, planning, and coding tasks, showing that information order, task structure, and memory compression shape contamination from invalidated hypotheses.
    \item We argue for uncertainty-preserving state management as a research agenda for interactive assistants: systems should track whether assumptions are tentative, confirmed, unresolved, or invalidated, rather than merely remembering their textual content.
\end{enumerate}

\section{Related Work}
\label{sec:related_work}

\paragraph{Dialogue degradation and path dependence.}
Recent work has shown that strong LLMs can degrade substantially in extended conversations, often by making early assumptions and failing to recover after taking a wrong turn~\citep{laban2025lost}. Other benchmarks evaluate dialogue reasoning and multi-step instruction following, showing that models struggle to integrate feedback across turns~\citep{li2025mtrbench,zhang2025turnbench}. Complementary benchmarks study structured flows and coding settings, where early mistakes can propagate into later artifacts~\citep{li2025structflowbench,rawal2025mtsec}. Our focus is narrower than overall dialogue degradation: we study the residue that remains after a wrong turn has been explicitly corrected. Rather than measuring only whether performance drops, we ask whether the final output still contains traces of an interpretation that the dialogue has invalidated.

\paragraph{Ambiguity, clarification, and underspecified intent.}
A parallel line of work studies ambiguity and clarification. Classical clarification work asks when a system should ask rather than answer~\citep{rao2018learning,aliannejadi2019asking}, while AmbigQA frames ambiguous questions as having multiple valid interpretations that must be made explicit~\citep{min2020ambigqa}. Recent LLM-focused benchmarks extend this problem to realistic dialogue settings: ClarifyMT-Bench reports an under-clarification bias, while AskBench evaluates when asking improves outcomes under underspecified queries~\citep{luo2025clarifymt,zhao2026askbench}. ClarEval and CLARITY further study clarification under ambiguous code and NL2SQL tasks~\citep{chen2026clareval,sarwar2026clarity}. Closest to our method, structured-uncertainty approaches use explicit uncertainty over tool parameters to decide which clarification question to ask~\citep{suri2025structured}. Our work is complementary: rather than only evaluating whether a model asks a question, we study what happens after the model has already committed and later receives a correction. The key failure is not only under-clarification, but failure to invalidate the stale state induced by the earlier interpretation.

\paragraph{Memory, summarization, and dialogue state.}
Long-context and retrieval-augmented systems aim to preserve information across extended interactions~\citep{lewis2020rag,liu2024lost}. Memory-augmented agents similarly rely on summaries, scratchpads, or external stores~\citep{packer2023memgpt,park2023generative}. Dialogue-state tracking has long treated interaction as belief-state update under uncertainty, especially in task-oriented systems with explicit slots~\citep{williams2007pomdp,young2013pomdp}. Our setting differs because the relevant latent state is open-ended: the ambiguous variable may be a tone, a planning constraint, a coding interface, or a hidden assumption about the user's goal. We therefore argue that memory should preserve epistemic status, not just content. A summary can remember what was said while forgetting whether it was tentative, contradicted, or unresolved; this is precisely how compression can induce posterior collapse.

\paragraph{Reasoning-time methods and posterior tracking.}
Inference-time reasoning methods make intermediate computation more explicit through chain-of-thought and self-consistency~\citep{wei2022cot,wang2023selfconsistency}. Related approaches use tree search, tool use, or reflection to revise intermediate trajectories~\citep{yao2023tot,shinn2023reflexion}. These methods are related to our posterior-tracking view because they often explore multiple reasoning trajectories. However, our results suggest that explicit reasoning or hypothesis enumeration is not enough. A model can list hypotheses, reason step by step, or revise locally while still carrying forward an invalidated task state. We therefore distinguish hypothesis enumeration from uncertainty-preserving inference: the latter requires maintaining tentative hypotheses, delaying commitment, invalidating contradicted assumptions, and regenerating from a revised belief state rather than patching the previous output.

\section{Theoretical Framing: Conversational Posterior Collapse}
\label{sec:theory}

A dialogue can be viewed as an incremental inference problem, echoing classical belief-state views of dialogue while extending them to open-ended LLM tasks~\citep{williams2007pomdp,henderson2014dstc}. At each turn, the assistant observes evidence from the user and must infer a latent task state: what the user wants, which constraints matter, which assumptions remain unresolved, and what output would satisfy the request. Let the dialogue history after receiving user turn $u_t$ be
\begin{equation}
    D_t = (u_1, a_1, \ldots, a_{t-1}, u_t),
\end{equation}
where $u_i$ denotes a user turn and $a_i$ denotes an assistant turn. Let $z \in \mathcal{Z}$ denote a latent task hypothesis, such as the intended audience of a writing request, the true objective of a planning task, or the intended API contract in a coding task. Ideally, the assistant should maintain a posterior distribution over hypotheses:
\begin{equation}
    p_t(z) = p(z \mid D_t).
\end{equation}

When the user request is ambiguous, this posterior should remain broad. If several task states are plausible, the assistant should preserve uncertainty rather than immediately select one. A later clarification $c$ should then update the posterior by Bayes' rule:
\begin{equation}
    p(z \mid D_t, c) \propto p(c \mid z, D_t) p_t(z).
\end{equation}

In an uncertainty-preserving assistant, hypotheses contradicted by the clarification should lose probability mass, while hypotheses supported by the clarification should gain mass. The assistant should commit to a final output only when the posterior is sufficiently concentrated or when the remaining uncertainty is irrelevant to the answer.

\paragraph{Definition.}
Let $Z$ be the set of plausible task hypotheses after an ambiguous dialogue prefix $D_t$, and let $c$ be a later clarification that supports a clarified hypothesis $z^+ \in Z$ while contradicting an earlier plausible hypothesis $z^- \in Z$. An uncertainty-preserving assistant should maintain uncertainty before $c$ and, after observing $c$, invalidate the contradicted hypothesis:
\begin{equation}
    p(z^- \mid D_t, c) \approx 0.
\end{equation}
We say \emph{early posterior collapse} occurs when the assistant behaves as if its task-state posterior has collapsed to $z^-$ before the ambiguity was resolved, and the later clarification fails to remove $z^-$ from the generation process:
\begin{equation}
    p_{\mathrm{eff}}(z^- \mid D_t, c) > 0.
\end{equation}
Here, $p_{\mathrm{eff}}$ denotes the model's behaviorally implied task-state distribution rather than a directly observed internal probability. We operationalize this residual effective probability through \emph{old-hypothesis contamination}: final outputs that retain lexical, semantic, or structural traces of $z^-$ despite the clarification.

Intuitively, early posterior collapse corresponds to the model behaving as if
\begin{equation}
    p_t(z) \approx \delta(z = z^-)
\end{equation}
before the dialogue evidence justifies such certainty. We use the term by analogy to probabilistic inference: it refers to premature commitment over latent task hypotheses in dialogue, not to variational posterior collapse in latent-variable models. The collapse need not appear as an explicit statement such as ``the user must mean $z^-$.'' It may appear behaviorally through examples, plans, code structure, tool choices, or response framing. The surface text may remain flexible while the hidden task state has already become overcommitted.

This gives a simple way to distinguish forgetting from early posterior collapse. In a forgetting failure, relevant evidence is missing from the effective context. In a posterior-collapse failure, relevant evidence is present, but the model fails to update the epistemic status of an earlier hypothesis. The model may verbally acknowledge that $z^-$ is no longer valid, while still generating an output shaped by $z^-$. This residual influence is what we operationalize as old-hypothesis contamination.

We can express contamination as the degree to which the final output $y$ is better explained by an invalidated hypothesis $z^-$ than by the clarified hypothesis $z^+$. Abstractly, let $\phi(y,z)$ measure the support that output $y$ gives to hypothesis $z$. Then a contamination score can be viewed as
\begin{equation}
    \mathcal{C}(y; z^-, z^+) = \max \left(0, \phi(y,z^-) - \phi(y,z^+) \right),
\end{equation}
or, more generally, as any evaluator that detects lexical, semantic, or structural traces of $z^-$ in $y$. The key point is that contamination is not equivalent to forgetting the clarification. It is evidence that the invalidated hypothesis remains active during generation.

This framing also explains why common memory mechanisms can fail, especially when compression preserves content but not uncertainty status~\citep{bai2023longbench,packer2023memgpt}. A summary can preserve facts while losing uncertainty status. Suppose the true posterior at turn $t$ is multi-modal:
\begin{equation}
    p_t(z_A) \approx p_t(z_B),
\end{equation}
because both interpretations are plausible. A lossy summary $s_t = S(D_t)$ may convert this into a single apparent state:
\begin{equation}
    p(z_A \mid s_t) \gg p(z_B \mid s_t),
\end{equation}
not because the dialogue resolved the ambiguity, but because the summarizer compressed it away. This is posterior collapse induced by memory compression.

Similarly, chain-of-thought and explicit state-tracking prompts do not automatically solve the problem, despite their value in exposing intermediate reasoning paths~\citep{wang2023selfconsistency}. Chain-of-thought may discuss alternatives without preserving calibrated uncertainty. A single-state ledger may make the dialogue state explicit while still forcing unresolved ambiguity into one settled representation. A true posterior-tracking system needs at least four operations: representing plausible hypotheses, tracking unresolved slots, updating with evidence, and invalidating contradicted states. Without these operations, explicit state representations can become additional anchors rather than safeguards.

This perspective leads to a practical design principle: interactive assistants should separate hypothesis management from answer generation. Before producing a committed answer, the model should identify unresolved slots, ask clarifying questions when necessary, and avoid encoding guesses as settled facts. After clarification, the model should update or reconstruct its task state rather than merely patch the previous output. The experiments below test whether current models behave in this uncertainty-preserving way.

\section{Diagnostic Evidence}
\label{sec:diagnostic_evidence}

Our diagnostic experiments test whether early posterior collapse explains recurring dialogue failures. The theory in Section~\ref{sec:theory} predicts three empirical signatures. First, model behavior should be path-dependent: the same final information can lead to different outcomes depending on when ambiguity is resolved. Second, tasks with stronger structural commitments should be more vulnerable, because early assumptions become embedded in the output. Third, dialogue-state representations such as summaries, chain-of-thought traces, or explicit state ledgers may fail when they preserve content without preserving epistemic status. We also include small supplementary interventions to test the corresponding design principle: if stale task state is the problem, then explicit rollback and two-phase uncertainty resolution should reduce old-hypothesis contamination.

\subsection{Experimental Setup}

Each task is a short dialogue in which the initial user request is under-specified and admits multiple plausible interpretations. A later turn resolves the ambiguity. The model must produce a final answer that satisfies the clarified intent while avoiding content from the earlier, invalidated hypothesis. We evaluate three task families: writing, planning, and coding. Writing tasks involve ambiguity in audience, tone, genre, or content target; planning tasks involve ambiguity in goals, constraints, priorities, or resources; coding tasks involve ambiguity in API contracts, data structures, implementation targets, or edge-case assumptions.

The main dataset contains 290 generated tasks across writing, planning, and coding, supplemented by 20 hand-crafted pilot tasks and a harder adversarial set of 30 hand-crafted coding and planning tasks. We evaluate Gemini-2.5-Pro and Gemini-2.5-Flash. Gemini-2.5-Pro is also used as the automatic judge with temperature 0. In total, the evaluation includes approximately 13,900 API calls and 7,160 trials.

We report task success, constraint satisfaction, revision behavior, wrong commitment, clarification behavior, and old-hypothesis contamination. The key metric is contamination: whether the final response retains lexical, semantic, or structural traces of an earlier interpretation that should have been invalidated by later clarification. A model can acknowledge the correction and still be contaminated if it preserves the old hypothesis in the final answer.

\subsection{Path Dependence: Same Information, Different Order}

If early posterior collapse is a state-commitment failure, then the path by which information is revealed should matter, not merely the final information available to the model. We therefore compare full, clarified-first, and ambiguous-first variants of the same tasks. The ambiguous-first condition gives the model an opportunity to form an early task state before clarification arrives.

\begin{table}[h]
\centering
\small
\begin{tabular}{llcc}
\toprule
\textbf{Model} & \textbf{Version} & \textbf{Success \%} & \textbf{Constraint \%} \\
\midrule
Gemini-2.5-Pro & Full & 54.2 & 74.3 \\
Gemini-2.5-Pro & Clarified-first & 50.0 & 74.2 \\
Gemini-2.5-Pro & Ambiguous-first & 42.8 & 74.1 \\
Gemini-2.5-Flash & Full & 62.3 & 77.9 \\
Gemini-2.5-Flash & Clarified-first & 55.4 & 75.7 \\
Gemini-2.5-Flash & Ambiguous-first & 45.7 & 71.8 \\
\bottomrule
\end{tabular}
\caption{Path dependence under equivalent final information. Ambiguous-first interactions reduce success for both models while leaving explicit constraint satisfaction comparatively stable.}
\label{tab:path_dependence}
\end{table}

As shown in Table~\ref{tab:path_dependence}, both models show lower success in the ambiguous-first condition than in the clarified-first condition, even though the final task-relevant information is equivalent. We treat this as an initial signature of path dependence rather than conclusive evidence by itself: order effects can overlap with recency bias, and contamination is not monotonic for every model in this experiment. The stronger mechanism test comes from the state-representation and intervention results below.

\subsection{Structural Embedding: Coding Is Most Vulnerable}

The posterior-collapse account predicts that tasks with stronger structural commitments should be more fragile. In writing, an early assumption may be locally edited. In coding, the same assumption can become embedded in interfaces, data schemas, dependencies, and control flow.

\begin{table}[h]
\centering
\small
\begin{tabularx}{\textwidth}{lccX}
\toprule
\textbf{Category} & \textbf{Success \%} & \textbf{Contamination} & \textbf{Interpretation} \\
\midrule
Writing & 69.3 & 0.125 & Easiest to locally revise. \\
Planning & 37.5 & 0.141 & Moderately vulnerable to stale constraints. \\
Coding & 24.4 & 0.256 & Most vulnerable; early assumptions become structural. \\
\bottomrule
\end{tabularx}
\caption{Task-family susceptibility in ambiguous-first interactions. Coding has the lowest success and highest contamination.}
\label{tab:structural_embedding}
\end{table}

Table~\ref{tab:structural_embedding} shows that coding has the lowest success and highest contamination, roughly doubling the contamination observed in writing. This supports the claim that early posterior collapse is not merely a linguistic anchoring effect. When the output has internal structure, early assumptions are harder to remove through local revision.

\subsection{Acknowledgement Is Not Revision}

We next test whether models that verbally acknowledge a correction actually revise their behavior. At the aggregate model level, the acknowledgement-behavior gap is modest and not always positive. However, the task-level breakdown reveals a sharper pattern.

\begin{table}[h]
\centering
\small
\begin{tabular}{lcccc}
\toprule
\textbf{Category} & \textbf{Ack \%} & \textbf{Revised \%} & \textbf{Gap \%} & \textbf{Contamination} \\
\midrule
Writing & 84.0 & 94.3 & -10.3 & 0.149 \\
Planning & 96.4 & 93.3 & +3.1 & 0.164 \\
Coding & 93.5 & 86.4 & +7.1 & 0.213 \\
\bottomrule
\end{tabular}
\caption{Acknowledgement-behavior gap by task family. Coding shows the largest positive gap: models acknowledge the correction more often than they behaviorally revise.}
\label{tab:ack_revision}
\end{table}

Table~\ref{tab:ack_revision} shows that coding has a +7.1 point acknowledgement-behavior gap. This suggests that the model can recognize the correction at the surface level while failing to rebuild the technical artifact around it. Acknowledgement is therefore not sufficient evidence of state revision.

\subsection{Memory and Reasoning Baselines Do Not Preserve Uncertainty}
We compare common strategies for maintaining or improving dialogue state: raw dialogue history, summary memory, chain-of-thought, and a single-state ledger. These baselines test whether the failure can be solved by more context, compressed memory, explicit reasoning, or forcing the model to maintain an explicit dialogue state.

As shown in Table~\ref{tab:baselines}, no baseline eliminates contamination. Summary memory
improves success relative to raw history but increases contamination, suggesting that summaries can preserve useful content while damaging epistemic status. Chain-of-thought does not reliably improve task success. The single-state ledger also increases contamination, consistent with the idea that forcing a single explicit state can prematurely settle ambiguity rather than preserve uncertainty.

\begin{table}[h]
\centering
\small
\begin{tabular}{lcccc}
\toprule
\textbf{Method} & \textbf{Success \%} & \textbf{Constraint \%} & \textbf{Contamination} & \textbf{Revised \%} \\
\midrule
Raw History & 43.2 & 72.5 & 0.177 & 90.7 \\
Summary Memory & 48.7 & 75.2 & 0.228 & 89.4 \\
Chain-of-Thought & 44.3 & 69.1 & 0.182 & 87.9 \\
Single-State Ledger & 44.6 & 71.9 & 0.295 & 85.8 \\
\bottomrule
\end{tabular}
\caption{Memory and reasoning baselines. Conventional strategies do not eliminate old-hypothesis contamination.}
\label{tab:baselines}
\end{table}

\subsection{State Representations Can Collapse or Distract}

We separately analyze two state representations: summaries and chain-of-thought traces. Summaries are intended to compress dialogue state, while chain-of-thought is intended to expose intermediate reasoning. Both can fail to preserve the epistemic status of assumptions.

\begin{table}[h]
\centering
\small
\begin{tabular}{lc}
\toprule
\textbf{Metric} & \textbf{Value} \\
\midrule
Summaries that preserve ambiguity & 37.5\% \\
Summaries that collapse uncertainty & 62.5\% \\
Contaminated by wrong hypothesis & 5.0\% \\
\bottomrule
\end{tabular}
\caption{Summary collapse analysis. Most summaries collapse ambiguity rather than preserving it.}
\label{tab:summary_collapse}
\end{table}

Table~\ref{tab:summary_collapse} shows that most summaries collapse uncertainty, converting unresolved ambiguity into apparent certainty. Chain-of-thought shows a different failure mode: it rarely makes explicit wrong commitments, but still underperforms in task success. This suggests that CoT may degrade performance through attentional diversion or over-analysis rather than overtly stating the wrong hypothesis.

\subsection{Interventions: Rollback and Two-Phase Policies}

As a supplementary mechanism check, if old-hypothesis contamination is caused by stale task state, then explicitly invalidating that state should help. We test this with a state rollback intervention: after clarification, the model is instructed to discard prior assumptions and rebuild from the clarified requirement.

\begin{table}[h]
\centering
\small
\begin{tabular}{lcc}
\toprule
\textbf{Condition} & \textbf{Success \%} & \textbf{Contamination} \\
\midrule
Normal clarification & 17.5 & 0.100 \\
Explicit rollback & 27.5 & 0.000 \\
\bottomrule
\end{tabular}
\caption{State rollback intervention. Explicit rollback improves success and eliminates measured contamination.}
\label{tab:rollback}
\end{table}

Table~\ref{tab:rollback} shows that explicit rollback improves success by 10 points and reduces measured contamination to zero in this supplementary setting. We do not interpret this as evidence that a prompt-level rollback instruction is a general solution. Rather, it supports the patching-vs-rebuilding account: ordinary clarification can behave like a local edit to an existing state, while rollback encourages reconstruction from the clarified requirement.

We also compare response policies that differ in when the model is allowed to commit.

\begin{table}[h]
\centering
\small
\begin{tabular}{lcc}
\toprule
\textbf{Policy} & \textbf{Committed Wrong \%} & \textbf{Contamination} \\
\midrule
Direct & 40.0 & 1.20 \\
List assumptions & 27.3 & 0.82 \\
Ask first & 0.0 & 0.40 \\
Two phase & 0.0 & 0.00 \\
\bottomrule
\end{tabular}
\caption{Response-policy comparison. The two-phase policy eliminates both wrong commitment and contamination in this setting.}
\label{tab:policy}
\end{table}

Table~\ref{tab:policy} suggests a hierarchy among these response policies in the small controlled setting. Listing assumptions helps but still permits execution under uncertainty. Asking first prevents explicit wrong commitment but can leave residual contamination. The two-phase policy, which resolves uncertainty before execution and then regenerates from the clarified state, eliminates both wrong commitment and measured contamination in this setting. 

\section{Discussion, Alternative Views, and Limitations}
\label{sec:discussion}
Our results suggest that dialogue robustness should be understood as a problem of epistemic state management, not only memory retention. A model can retain the text of a clarification and still fail if an earlier ambiguous interpretation has already become the active task state. This explains why longer context or compressed memory alone may be insufficient: the system must preserve not only content, but also whether each assumption is confirmed, tentative, contradicted, or unresolved.

Our diagnostic experiments do not fully separate stale-state contamination from all forms of order sensitivity, including recency bias, prompt-position effects, or instruction-following brittleness. The position of this paper is therefore not that old-hypothesis contamination is the only explanation for dialogue degradation, but that it is a distinct and under-measured residue of failed state invalidation. Future benchmarks should separately measure whether a model remembers a clarification, acknowledges it, and actually removes invalidated assumptions from the generated artifact.

The evidence in Table~\ref{tab:baselines} also clarifies the role of explicit state representations. Making reasoning or state more visible is not the same as preserving uncertainty. Chain-of-thought can expose intermediate reasoning without ensuring revision, while a single-state ledger can make the dialogue state explicit but still collapse unresolved ambiguity into a settled representation. The missing ingredient is the ability to update, arbitrate, and invalidate assumptions as new evidence arrives. In this sense, posterior tracking is a state-management problem, not just a prompting pattern.

A natural alternative view is that the observed failures are simply order effects, recency bias, or instruction-following brittleness. We do not rule out these explanations. Instead, our position is that old-hypothesis contamination captures a distinct residue that existing evaluations often miss: whether the model removes an interpretation that later dialogue has made obsolete.

Our study has several limitations. We evaluate two Gemini models on a controlled task distribution, and use a model-based judge, so the results should be replicated across additional model families and with human evaluation. Our contamination metric is also imperfect: old-hypothesis traces can be lexical, semantic, or structural, and structural contamination is especially difficult to measure in coding tasks. Finally, the state-management agenda is supported here only by prompt-level approximations, such as rollback and two-phase response policies. The rollback and two-phase experiments should also be interpreted cautiously: they are small, prompt-level mechanism checks, not evidence that a simple prompt solves the general problem. Their role is to distinguish patching from rebuilding and to motivate more explicit state-management mechanisms. A full implementation may require explicit dialogue-state controllers, memory schemas that preserve epistemic status, or training objectives that reward uncertainty preservation; Appendix~\ref{app:upi} sketches one possible direction.

Our future work should also test whether the same failure mode is amplified in tool-using or memory-augmented agents, where early assumptions can propagate into external state and require rollback.

\section{Conclusion}
\label{sec:conclusion}
Clarification is not always correction. A model may receive the right evidence, acknowledge it, and still produce an answer shaped by an interpretation that should have been invalidated. This paper argues that such old-hypothesis contamination is a useful diagnostic target for LLMs in dialogue. The central challenge is not only whether an assistant remembers what was said, but whether it maintains the epistemic status of what was said: which assumptions remain tentative, which have been confirmed, and which must be discarded. Robust interactive systems should therefore be evaluated not only for memory retention or final-task success, but also for their ability to let go of stale hypotheses after clarification.
\newpage
\bibliographystyle{unsrtnat}
\bibliography{references}
\appendix

\section{Sketch of Uncertainty-Preserving State Management}
\label{app:upi}

The diagnostic evidence in the main paper suggests a simple design principle: if the failure is early posterior collapse, the remedy is to delay collapse until ambiguity has been resolved and to explicitly invalidate stale task states when later evidence contradicts them. Each component of this principle is motivated by an empirical result. The order-sensitivity experiment shows that early ambiguity can shape later behavior. The summary, ledger, and supplementary multi-hypothesis experiments show that representing dialogue state is not enough unless uncertainty is preserved and contradicted hypotheses can be removed. The rollback experiment shows that ordinary clarification is often treated as a patch, while explicit state invalidation reduces contamination. The response-policy experiment shows that separating uncertainty resolution from task execution can eliminate wrong commitment and contamination in our supplementary setting.

We sketch \emph{uncertainty-preserving inference} as one possible implementation of this agenda for interactive assistants. This is not intended as a new model architecture. Instead, it is a system-level inference policy that can be implemented through prompting, agent controllers, memory schemas, or dialogue-state logic. The goal is to make the assistant track not only what has been said, but also the epistemic status of what has been said: which assumptions are confirmed, which remain tentative, and which have been invalidated by later evidence.

The protocol separates response generation into two phases. The first phase manages uncertainty; the second phase executes the task. Let
\[
H_t = \{z_1, \ldots, z_k\}
\]
be the set of plausible task hypotheses after observing dialogue \(D_t\). Let \(S_t\) be the set of unresolved slots whose values would materially change the final answer:
\[
S_t = \{s : \mathrm{Impact}(s, y) > \tau\}.
\]
Here, \(\mathrm{Impact}(s, y)\) denotes how much the answer \(y\) would change depending on the value of slot \(s\), and \(\tau\) is a task-dependent threshold. If \(S_t\) is non-empty, the model should not produce a committed answer. Instead, it should ask a clarification question targeted at the highest-impact unresolved slots.

The decision rule is:
\[
\mathrm{Act}(D_t) =
\begin{cases}
\mathrm{Clarify}(S_t), & \text{if } |S_t| > 0, \\
\mathrm{Execute}(z^\ast), & \text{otherwise},
\end{cases}
\quad
\text{where }
z^\ast = \arg\max_z p_t(z).
\]
This rule operationalizes the central claim of the paper: the assistant should not collapse its posterior into a committed output while high-impact uncertainty remains unresolved.

When clarification arrives, the protocol performs explicit hypothesis invalidation. Let \(c\) be the clarification and let \(I(c,z)\) indicate whether hypothesis \(z\) is inconsistent with the clarification. The updated hypothesis set is:
\[
H_{t+1}
=
\{z \in H_t : I(c,z)=0\}
\cup H_{\mathrm{new}}(c).
\]
The model should then regenerate from the revised hypothesis state rather than patching the previous output:
\[
y_{t+1}
=
G(H_{t+1}, D_t, c),
\quad
\text{not}
\quad
y_{t+1}
=
\mathrm{Patch}(y_t, c).
\]
This distinction directly follows from the rollback experiment. A patching operation modifies the earlier answer while preserving much of its latent structure. A rebuilding operation generates from a revised belief state in which contradicted assumptions have been removed. In our experiments, explicitly inducing this rebuilding behavior improves success and eliminates measured contamination.

The protocol can be summarized as in Table \ref{tab:uncertainty_preserving_inference}, and pseudocode as in Alg. \ref{alg:upi}. In the pseudocode, $H_t$ denotes candidate task hypotheses and $S_t$ denotes high-impact unresolved slots. The key operation is \textsc{InvalidateAndRebuild}: when clarification contradicts an earlier assumption, the system removes the stale hypothesis and regenerates from the revised belief state rather than patching the previous output.

\begin{table}[h]
\centering
\small
\begin{tabularx}{\textwidth}{p{0.24\textwidth} p{0.36\textwidth} X}
\toprule
\textbf{Stage} & \textbf{Operation} & \textbf{Empirical Motivation} \\
\midrule
Ambiguity detection
& Identify unresolved slots \(S_t\).
& Order sensitivity shows that unresolved ambiguity can shape later behavior. \\

Hypothesis tracking
& Maintain plausible task states \(H_t\) as tentative.
& Naive single-state representations can collapse uncertainty too early. \\

Clarification decision
& Ask when unresolved slots affect the answer.
& Clarification policies prevent premature wrong commitment. \\

Hypothesis invalidation
& Remove states contradicted by clarification.
& Rollback reduces contamination by explicitly invalidating stale assumptions. \\

Regeneration
& Generate from revised beliefs, not patched output.
& Two-phase policies succeed when execution follows uncertainty resolution. \\
\bottomrule
\end{tabularx}
\caption{Uncertainty-preserving inference protocol and its empirical motivation.}
\label{tab:uncertainty_preserving_inference}
\end{table}

\begin{algorithm}[h]
\caption{Uncertainty-Preserving Inference}
\label{alg:upi}
\begin{algorithmic}[1]
\Require Dialogue history $D_t$, belief state $B_t$
\State $H_t, S_t \leftarrow \textsc{UpdateState}(B_t, D_t)$

\If{$S_t \neq \emptyset$}
    \State \Return \textsc{Clarify}$(S_t)$
\EndIf

\State $z^\ast \leftarrow \textsc{SelectConfirmed}(H_t)$
\State \Return \textsc{Generate}$(z^\ast, B_t)$

\Statex
\Procedure{OnClarification}{$c, B_t$}
    \State $B_{t+1} \leftarrow \textsc{InvalidateAndRebuild}(B_t, c)$
    \State \Return \textsc{Generate}$(B_{t+1})$
\EndProcedure
\end{algorithmic}
\end{algorithm}

The response-policy experiment in Table~\ref{tab:policy} can be interpreted as a prompt-level approximation of this protocol. Direct answering collapses immediately. Listing assumptions externalizes uncertainty but still permits execution under uncertainty. Ask-first delays commitment but does not necessarily enforce state rebuilding. The two-phase policy most closely matches uncertainty-preserving inference: it first resolves unresolved slots, then executes from the clarified state. Its elimination of both wrong commitment and contamination provides initial evidence that the protocol addresses the failure mode identified in this paper.

This protocol should be understood as a design primitive rather than a complete solution. The experiments do not show that prompt-level uncertainty preservation is sufficient for all dialogue settings. Instead, they show that the core operations of the protocol---delaying commitment, invalidating contradicted hypotheses, and regenerating from revised state---directly target the mechanisms exposed by our experiments. More robust implementations may require explicit dialogue-state controllers, memory schemas that store epistemic status, or model architectures trained to maintain and revise uncertainty over user intent.

\section{Additional Experimental Details}
\label{app:experimental_details}

This appendix provides additional details for the experimental setup, task construction, evaluation metrics, and secondary result tables. The main paper organizes experiments by conceptual claims: path dependence, structural embedding, state-representation failure, and intervention. Here we list the full experiment suite explicitly and include supporting tables that are useful for reproducibility but too detailed for the main text.

\subsection{Task Families}

All tasks are constructed as short dialogues. The first user turn is intentionally under-specified and admits more than one plausible interpretation. A later turn resolves the ambiguity. The model is then evaluated on whether its final answer follows the clarified intent and avoids traces of the earlier, invalidated interpretation.

We use three task families. Writing tasks test whether models can revise audience, tone, genre, or content framing. Planning tasks test whether models can update goals, constraints, priorities, or resource assumptions. Coding tasks test whether models can revise structural assumptions such as API contracts, data structures, implementation targets, or edge-case behavior. Coding is especially important for our hypothesis because early assumptions can become embedded in the internal structure of the output, not merely in surface wording.

\begin{table}[h]
\centering
\small
\begin{tabularx}{\textwidth}{p{0.16\textwidth} p{0.30\textwidth} X}
\toprule
\textbf{Task Family} & \textbf{Ambiguity Type} & \textbf{Example Failure Mode} \\
\midrule
Writing & Audience, tone, genre, or content target & The model keeps the original audience framing after the user changes it. \\
Planning & Goal, constraints, priorities, or resource assumptions & The model preserves an earlier travel, budget, or scheduling assumption. \\
Coding & API contract, data structure, implementation target, or edge-case assumption & The model keeps an earlier interface or control-flow assumption after clarification. \\
\bottomrule
\end{tabularx}
\caption{Task families and ambiguity types.}
\label{tab:app_task_families}
\end{table}

\subsection{Datasets and Run Scale}

The full dataset contains 290 generated tasks across writing, planning, and coding. We also use 20 hand-crafted pilot tasks and a harder adversarial validation set of 30 hand-crafted coding and planning tasks. The generated tasks provide coverage and statistical scale, while the hand-crafted tasks are designed to stress-test the phenomenon under more adversarial ambiguity.

The full run uses Gemini-2.5-Pro and Gemini-2.5-Flash as evaluated models. Gemini-2.5-Pro is also used as the automatic judge with temperature 0. The complete evaluation includes approximately 13,900 API calls and 7,160 total trials, including 6,960 pilot/full-dataset trials and 200 bonus/intervention trials.

\begin{table}[h]
\centering
\small
\begin{tabular}{ll}
\toprule
\textbf{Quantity} & \textbf{Value} \\
\midrule
Total API calls & $\sim$13,900 \\
Total trials & 6,960 + 200 bonus trials \\
Generated tasks & 290 \\
Hand-crafted pilot tasks & 20 \\
Adversarial validation tasks & 30 \\
Task categories & Writing, Planning, Coding, Mixed \\
Evaluated models & Gemini-2.5-Pro, Gemini-2.5-Flash \\
Judge & Gemini-2.5-Pro, temperature 0 \\
\bottomrule
\end{tabular}
\caption{Evaluation scale and dataset composition.}
\label{tab:app_dataset_scale}
\end{table}

\subsection{Evaluation Metrics}

Table~\ref{tab:app_metrics} defines the metrics used throughout the evaluation. The key metric is old-hypothesis contamination. This metric captures whether the final response retains lexical, semantic, or structural traces of an earlier interpretation that should have been invalidated by later clarification.

This distinction is important because a model can verbally acknowledge a correction while still producing a contaminated answer. For example, a coding response may say it understands the corrected API requirement but still preserve a function signature from the earlier hypothesis. Similarly, a writing response may acknowledge a changed audience but retain the tone or framing of the original audience.

\begin{table}[h]
\centering
\small
\begin{tabularx}{\textwidth}{lX}
\toprule
\textbf{Metric} & \textbf{Definition} \\
\midrule
Success & Whether the final response satisfies the clarified user request. \\
Constraint Satisfaction & Whether explicit constraints in the final clarified request are followed. \\
Revised & Whether the model behaviorally updates after clarification. \\
Acknowledged & Whether the model verbally acknowledges the clarification. \\
Ack-Behavior Gap & Difference between verbal acknowledgement and actual behavioral revision. \\
Contamination & Degree to which the final response retains content from the earlier, incorrect hypothesis. \\
Committed Wrong & Whether the model explicitly commits to an incorrect hypothesis before ambiguity is resolved. \\
Asked Clarification & Whether the model asks a clarifying question instead of prematurely answering. \\
\bottomrule
\end{tabularx}
\caption{Evaluation metrics.}
\label{tab:app_metrics}
\end{table}

\subsection{Complete Experiment Suite}

The main paper presents experiments by conceptual role rather than by internal experiment ID. Table~\ref{tab:app_experiment_suite} lists the full experiment suite for completeness. The experiments are designed to distinguish three explanations for dialogue failure: memory loss, insufficient reasoning, and premature state commitment.

The first group tests signatures of premature commitment, such as order sensitivity, acknowledgement-behavior gaps, and task-family susceptibility. The second group evaluates whether common memory or reasoning strategies preserve uncertainty. The third group tests mechanism and intervention: if stale task state is the problem, then explicit rollback and two-phase response policies should reduce contamination.

\begin{table}[h]
\centering
\small
\begin{tabularx}{\textwidth}{llX}
\toprule
\textbf{ID} & \textbf{Experiment} & \textbf{Purpose} \\
\midrule
Exp 1 & Same Information, Different Order & Tests whether final behavior changes when equivalent information appears in different orders. \\
Exp 2 & Acknowledgement-Behavior Gap & Tests whether verbal acknowledgement predicts actual revision. \\
Exp 3 & Old-Hypothesis Contamination & Measures contamination across prompting and memory methods. \\
Exp 4 & Baselines Comparison &Compares raw history, summary memory, CoT, and single-state ledger. \\
Exp 5 & K-Hypothesis Ablation & Tests whether maintaining more hypotheses improves robustness. \\
Exp 6 & Summary Collapse & Tests whether summaries preserve or collapse uncertainty. \\
Exp 7 & CoT Commitment Analysis & Tests whether CoT explicitly commits to wrong hypotheses. \\
Exp 8 & Category Susceptibility & Compares contamination across writing, planning, and coding. \\
Exp 9 & Clarification Policy & Tests whether ask-when-uncertain policies prevent wrong commitment. \\
Exp 10 & State Rollback & Tests whether explicit state invalidation improves revision. \\
Exp 11 & Response Policy Comparison & Compares direct answering, assumption listing, ask-first, and two-phase policies. \\
Exp 12 & Adversarial Validation & Tests the phenomenon on hand-crafted hard coding/planning tasks. \\
\bottomrule
\end{tabularx}
\caption{Complete experiment suite.}
\label{tab:app_experiment_suite}
\end{table}

\section{Additional Results}

\subsection{Model-Level Acknowledgement-Behavior Gap}

The main paper reports the category-level acknowledgement-behavior gap because the task-family breakdown more clearly shows where acknowledgement fails to imply revision. Table~\ref{tab:app_ack_model} reports the aggregate model-level results.

At the model level, the gap is modest. Gemini-2.5-Pro acknowledges corrections slightly more often than it behaviorally revises, while Gemini-2.5-Flash revises slightly more often than it explicitly acknowledges. This is why we do not claim that acknowledgement-behavior mismatch is universal. Instead, the stronger result is task-specific: the gap is most visible in coding tasks, where early assumptions become structural.

\begin{table}[h]
\centering
\small
\begin{tabular}{lcccc}
\toprule
\textbf{Model} & \textbf{Ack \%} & \textbf{Revised \%} & \textbf{Gap \%} & \textbf{Contamination} \\
\midrule
Gemini-2.5-Pro & 92.9 & 90.5 & +2.5 & 0.177 \\
Gemini-2.5-Flash & 89.5 & 92.7 & -3.3 & 0.171 \\
\bottomrule
\end{tabular}
\caption{Model-level acknowledgement-behavior gap.}
\label{tab:app_ack_model}
\end{table}

\subsection{Old-Hypothesis Contamination Across Methods}

For completeness, this appendix also reports the naive multi-hypothesis baseline that is omitted from the main baseline table. We treat this result as a supplementary negative control rather than as a primary baseline. Table~\ref{tab:app_contamination_methods} reports the contamination-only analysis across prompting and memory methods.

The key pattern is that none of the methods eliminates contamination. Summary memory increases mean contamination relative to raw history, suggesting that summarization can compress away epistemic status. Single-state ledger also increases contamination, consistent with the idea that forcing a single explicit state can prematurely settle ambiguity. Naive multi-hypothesis prompting produces the highest contamination, suggesting that listing hypotheses without arbitration or invalidation can preserve stale hypotheses rather than remove them.

\begin{table}[h]
\centering
\small
\begin{tabular}{lcc}
\toprule
\textbf{Method} & \textbf{Mean Contamination} & \textbf{Score $\geq$ 2 (\%)} \\
\midrule
Raw History & 0.177 & 1.2 \\
Summary Memory & 0.228 & --- \\
Chain-of-Thought & 0.182 & 2.5 \\
Single-State Ledger & 0.295 & --- \\
Multi-Hypothesis & 1.029 & 11.2 \\
\bottomrule
\end{tabular}
\caption{Old-hypothesis contamination across prompting and memory methods (Dashes indicate that this thresholded statistic was not computed for that condition in the original run).}
\label{tab:app_contamination_methods}
\end{table}

\subsection{Per-Model K-Hypothesis Ablation}

Although we omit naive multi-hypothesis prompting from the main baseline table for clarity, we include this supplementary ablation because it illustrates an important negative result: enumerating more hypotheses does not by itself implement posterior tracking. Table~\ref{tab:app_k_model} reports the per-model breakdown.

For Gemini-2.5-Pro, increasing the number of maintained hypotheses improves success from 4.7\% at $K=1$ to 19.1\% at $K=5$. For Gemini-2.5-Flash, the trend reverses: success drops from 22.6\% at $K=1$ to 5.0\% at $K=5$. This instability suggests that hypothesis enumeration is not the same as posterior tracking. More hypotheses can help only if the system can update, arbitrate, and invalidate them.

\begin{table}[h]
\centering
\small
\begin{tabular}{lccc}
\toprule
\textbf{Model} & \textbf{$K=1$ Success \%} & \textbf{$K=3$ Success \%} & \textbf{$K=5$ Success \%} \\
\midrule
Gemini-2.5-Pro & 4.7 & 10.7 & 19.1 \\
Gemini-2.5-Flash & 22.6 & 5.8 & 5.0 \\
\bottomrule
\end{tabular}
\caption{Per-model success by number of maintained hypotheses.}
\label{tab:app_k_model}
\end{table}

\subsection{Chain-of-Thought Commitment Analysis}

Table~\ref{tab:app_cot_commitment} reports the detailed chain-of-thought commitment analysis. We measure whether the reasoning trace explicitly commits to the wrong hypothesis, whether it later revises that commitment, whether final output remains contaminated, and the mean strength of wrong commitment.

The result suggests that CoT does not usually fail by openly stating the wrong hypothesis. Wrong commitment during reasoning is relatively rare, final explicit contamination is zero in this analysis, and the mean commitment strength is low. However, CoT still underperforms in the broader baseline comparison. This supports the interpretation that CoT can hurt through attentional diversion or over-analysis rather than through explicit wrong assertion.

\begin{table}[h]
\centering
\small
\begin{tabular}{lc}
\toprule
\textbf{Metric} & \textbf{Value} \\
\midrule
Commits to wrong hypothesis during reasoning & 7.5\% \\
Revises after commitment & 20.0\% \\
Final output contaminated & 0.0\% \\
Mean commitment strength, 1--5 & 1.35 \\
\bottomrule
\end{tabular}
\caption{Chain-of-thought commitment analysis.}
\label{tab:app_cot_commitment}
\end{table}

\subsection{Clarification Policy}

Table~\ref{tab:app_clarification_policy} reports the clarification-policy experiment. This experiment isolates a simple intervention: instructing the model to ask a clarification question when uncertainty remains.

The direct policy rarely asks clarification and commits to a wrong hypothesis 42.5\% of the time. The clarification policy asks in all cases and reduces wrong commitment to zero. This result supports the central claim that many failures occur before the model has enough evidence to answer. It also motivates the two-phase policy in the main paper: asking is useful, but the strongest intervention also requires executing from the clarified state rather than patching the earlier one.

\begin{table}[h]
\centering
\small
\begin{tabular}{lccc}
\toprule
\textbf{Policy} & \textbf{Asked Clarification \%} & \textbf{Committed Wrong \%} & \textbf{Would Help \%} \\
\midrule
Direct & 15.0 & 42.5 & 100.0 \\
Clarification & 100.0 & 0.0 & 97.5 \\
\bottomrule
\end{tabular}
\caption{Clarification policy experiment.}
\label{tab:app_clarification_policy}
\end{table}

\subsection{State Rollback and Response-Policy Interventions}

The intervention experiments test the mechanism most directly. If old-hypothesis contamination is caused by stale task state, then explicitly invalidating the earlier state should reduce contamination. Similarly, if premature commitment is the problem, then a policy that separates uncertainty resolution from task execution should outperform direct answering.

Table~\ref{tab:app_intervention_summary} summarizes the intervention and validation experiments. State rollback improves success and eliminates contamination, indicating that ordinary clarification does not always invalidate stale assumptions. The response-policy comparison shows that two-phase interaction eliminates both wrong commitment and contamination in the supplementary setting. The adversarial validation experiment shows that contamination remains observable even when tasks are deliberately difficult.

\begin{table}[h]
\centering
\small
\begin{tabularx}{\textwidth}{lXX}
\toprule
\textbf{Experiment} & \textbf{Main Finding} & \textbf{Interpretation} \\
\midrule
State Rollback & Rollback improves success and eliminates contamination. & Prior assumptions persist unless explicitly invalidated. \\
Response Policies & Two-phase policy eliminates wrong commitment and contamination. & Robustness requires separating uncertainty resolution from execution. \\
Adversarial Validation & Hard tasks show floor-level success but preserve contamination direction. & Contamination is observable even under adversarial ambiguity. \\
\bottomrule
\end{tabularx}
\caption{Summary of intervention and validation experiments.}
\label{tab:app_intervention_summary}
\end{table}

\end{document}